\documentclass{article}

\usepackage{arxiv}

\usepackage[utf8]{inputenc} 
\usepackage[T1]{fontenc}    
\usepackage{hyperref}       
\usepackage{url}            
\usepackage{booktabs}       
\usepackage{amsfonts}       
\usepackage{nicefrac}       
\usepackage{microtype}      
\usepackage{cleveref}       
\usepackage{lipsum}         
\usepackage{graphicx}
\usepackage{natbib}
\usepackage{doi}
\usepackage{multirow}

\title{DeepMedcast: A Deep Learning Method for Generating Intermediate Weather Forecasts among Multiple NWP Models}

\author{ \href{https://orcid.org/0000-0000-0000-0000}{\hspace{1mm}Atsushi Kudo} \\
	Numerical Prediction Division \\
	Japan Meteorological Agency \\
	Tokyo, Japan \\
	\texttt{atsushi.kudou@met.kishou.go.jp } \\
}

\renewcommand{\shorttitle}{DeepMedcast: Generating Intermediate Weather Forecasts}

\hypersetup{
pdftitle={DeepMedcast: Generating Intermediate Weather Forecasts among Multiple NWP Models Using Deep Neural Networks},
pdfsubject={},
pdfauthor={Atsushi Kudo},
pdfkeywords={Deep Neural network, Intemediate forecast, Numerical weather prediction},
}

\begin{document}
\maketitle

\begin{abstract}
Numerical weather prediction (NWP) centers around the world operate a variety of NWP models. In addition, recent advances in AI-driven NWP models have further increased the availability of NWP outputs. While this expansion holds the potential to improve forecast accuracy, it raises a critical question: which prediction is the most plausible? If the NWP models have comparable accuracy, it is impossible to determine in advance which one is the best. Traditional approaches, such as ensemble or weighted averaging, combine multiple NWP outputs to produce a single forecast with improved accuracy. However, they often result in meteorologically unrealistic and uninterpretable outputs, such as the splitting of tropical cyclone centers or frontal boundaries into multiple distinct systems.

To address this issue, we propose DeepMedcast, a deep learning method that generates intermediate forecasts between two or more NWP outputs. Unlike averaging, DeepMedcast provides predictions in which meteorologically significant features—such as the locations of tropical cyclones, extratropical cyclones, fronts, and shear lines—approximately align with the arithmetic mean of the corresponding features predicted by the input NWP models, without distorting meteorological structures. We demonstrate the capability of DeepMedcast through case studies and verification results, showing that it produces realistic and interpretable forecasts with higher accuracy than the input NWP models. By providing plausible intermediate forecasts, DeepMedcast can significantly contribute to the efficiency and standardization of operational forecasting tasks, including general, marine, and aviation forecasts.
\end{abstract}


\section{Introduction}
\label{sec:sec1}
In recent decades, numerical weather predictions (NWPs) and their post-processing have played a central role in issuing weather forecasts, warnings, and advisories \citep{WMO2013, Vannitsem2021}. NWP centers around the world have developed and are operating a variety of NWP models for accurate weather predictions. For example, the European Centre for Medium-Range Weather Forecasts (ECMWF) operates the Integrated Forecasting System (IFS) and its ensemble prediction system \citep{ECMWF2024}; the UK Met Office operates the Unified Model and the Met Office Global and Regional Ensemble Prediction System \citep{Brown2012, Hagelin2017, Inverarity2023}. The National Centers for Environmental Prediction (NCEP) at the National Oceanic and Atmospheric Administration (NOAA) operates the Global Forecast System \citep{NCEP2016}, the High-Resolution Rapid Refresh \citep{Dowell2022}, and the Hurricane Weather Research and Forecasting model \citep{Gopalakrishnan2011}. The Japan Meteorological Agency (JMA) operates three deterministic NWP models and two ensemble prediction systems for short-range to weekly forecasts: the Global Spectrum Model (GSM), the Meso-Scale Model (MSM), the Local Forecast Model, the Global Ensemble Prediction System, and the Mesoscale Ensemble Prediction System \citep{JMA2024}. These models cover different areas with varying resolutions and processes.

In addition to traditional physics-based NWP models, recent advancements in artificial intelligence (AI) have introduced new methods for producing weather predictions. AI-driven NWP models, such as FourCastNet \citep{Pathak2022, Bonev2023}, GraphCast \citep{Lam2022}, Pangu-Weather \citep{Bi2022, Bi2023}, FengWu \citep{Chen2023, Han2024}, Aurora \citep{Bodnar2024}, GenCast \citep{Price2023}, and AIFS \citep{Lang2024}, have demonstrated the ability to enhance both the speed and accuracy of weather predictions by leveraging deep learning techniques to model complex atmospheric systems.

At present, forecasters are able to use multiple NWP models including AI-driven NWP models, which provide a range of possible atmospheric states, allowing them to select the most plausible prediction from available NWPs. However, this raises a critical question: Which prediction is the most plausible? If the models have comparable accuracy, it is impossible to determine in advance which one is the best. One practical and widely used solution is to average the results from multiple NWP models or their post-processed outputs, as this can reduce random errors inherent in NWP models and can lead to higher accuracy than individual models \citep{Vislocky1997, JMA2018}. The National Hurricane Center and the Joint Typhoon Warning Center in the United States use consensus forecasts (e.g., \cite{Simon2018, Cangialosi2023}), which are weighted averages, extensively for both tropical cyclone (TC) track and intensity predictions. JMA employs consensus forecasting for TC track predictions by averaging positions of TC centers from multiple NWP outputs to improve forecast accuracy \citep{Nishimura2019, JMA2022}. The UK Met Office operates the IMPROVER system \citep{Roberts2023}, which applies a weighted average of post-processing and nowcasts based on multiple NWP outputs. The National Weather Service (NWS) at NOAA operates the National Blend of Models (NBM), which provides statistically post-processed multi-model ensemble guidance \citep{Hamill2017}. The German Meteorological Service uses MOSMIX and ModelMIX \citep{Primo2024}, which are weighted averages of post-processing based on IFS, their global model, and their regional ensemble model. Additionally, the World Area Forecast Centre, comprising centers in London and Washington, operates harmonized forecasts, including mean, maximum, and minimum forecasts, from both NWP outputs for aviation hazards such as cumulonimbus clouds, turbulence, and in-flight icing \citep{ICAO2016}.

It is straightforward to average the central position of TCs, extra tropical cyclones, or the location of fronts because averaging does not degrade their clarity. However, averaging atmospheric fields such as pressure or wind speed around these systems is not appropriate. This is because averaging can smooth out or distort these fields, weakening the central pressure or wind speeds around cyclones and fronts, and even causing TCs or fronts to split into two, resulting in predictions that are meteorologically unrealistic and difficult to interpret. Forecasters must then choose between two options: using a single model that is realistic and interpretable but potentially less accurate or using an averaged prediction that is unrealistic and uninterpretable but may be more accurate.

Beyond these challenges, weather forecasting faces additional difficulties in operational practice. In JMA’s forecasting and warning issuance process, TC track forecasts, which are based on consensus forecasts derived by averaging the positions of TC centers from multiple NWP models, take precedence. As a result, forecasters responsible for general, marine, and aviation forecasts must ensure that their forecasts align with TC track forecasts. However, since no NWP model inherently conforms to the TC track forecasts, forecasters need to adjust the existing NWP outputs in their minds to construct forecast scenarios that align with them. This process requires significant time and effort and can pose a major obstacle to the standardization of forecasting workflows, leading to inefficiencies in operational forecasting.

In addition to these operational challenges, machine learning-based post-processing presents its own set of difficulties, particularly regarding data requirements. In conventional model output statistics (MOS), obtaining long-term, homogeneous datasets is particularly difficult because the input NWP model is periodically updated, causing changes in its systematic errors. Consequently, the statistical relationships learned from past data may no longer hold after a model update.

The objective of this study is to propose DeepMedcast, a method that uses deep learning to generate a realistic and interpretable "intermediate forecast" between two or more NWP models. In this study, we do not attempt to define intermediate forecasts in a physical or mathematical sense. Instead, we adopt a pragmatic definition: an intermediate forecast is a predicted meteorological field in which meteorologically significant features—such as the center positions of TCs or extratropical cyclones, frontal boundaries, and shear lines—are located at the arithmetic mean of the corresponding features predicted by input NWP models, and it simultaneously exhibits higher forecast accuracy against observations than the input models.

Unlike averaging, DeepMedcast can produce atmospheric fields around cyclones and fronts without smoothing out or disturbing their distributions. This capability is crucial in operational forecasts, where accurate and interpretable predictions are needed for issuing reliable warnings and advisories—particularly when a TC is approaching—and also contributes to the standardization of forecasting workflows by reducing reliance on manual adjustments by individual forecasters. While mathematical frameworks such as displacement interpolation and barycenters in optimal transport theory \citep{McCann1997, Agueh2011, Peyre2020} provide rigorously defined intermediate states and have been widely used in machine learning fields such as image processing, applying them directly to forecasts from multiple NWP models remains challenging. \cite{LeCoz2023} proposed a barycenter-based method using optimal transport to combine forecasts from multiple NWP models in the context of subseasonal prediction, offering a mathematically principled approach to multi-model ensemble forecasts. \cite{Duc2024} pointed out that traditional arithmetic ensemble means tend to excessively smooth rainfall structures, and proposed a Gaussian–Hellinger barycenter based on unbalanced optimal transport theory to derive more realistic and structurally coherent ensemble means. Their method is particularly effective in representing heavy precipitation and may contribute to future advances in ensemble post-processing and spatial verification techniques. To the best of the authors’ knowledge, however, no prior study has successfully generated two-dimensional intermediate forecasts for mean sea-level pressure or surface wind vectors at high temporal resolution and in areas strongly influenced by topography. In contrast, DeepMedcast is not intended to generate physically or mathematically consistent intermediate forecasts but to provide forecasters with operationally practical solutions, addressing a critical challenge in operational forecasting. In many national meteorological centers, including JMA, machine learning-based post-processing methods, referred to as forecast guidance, are operationally employed. Although forecast guidance does not preserve physical consistency, it enhances forecast accuracy by reducing biases inherent in NWP models and significantly improves the efficiency of operational forecasting. DeepMedcast is a kind of post-processing method that is not designed to ensure physical consistency but to provide practical support for forecasters.

This paper is structured as follows: Section 2 presents the methodology and data used for DeepMedcast, detailing the deep learning architecture and training process. Section 3 discusses the results of applying DeepMedcast to multiple NWP models with case studies and verification results, and Section 4 offers a discussion of contributions to operational forecasting and the key features of DeepMedcast's architecture. Finally, Section 5 concludes with a summary of the findings and future research directions.

\section{Method and data}
\label{sec:sec2}

\begin{figure}[b]
  \centering
  \includegraphics[width=0.70\textwidth,trim=0 0 0 0,clip]{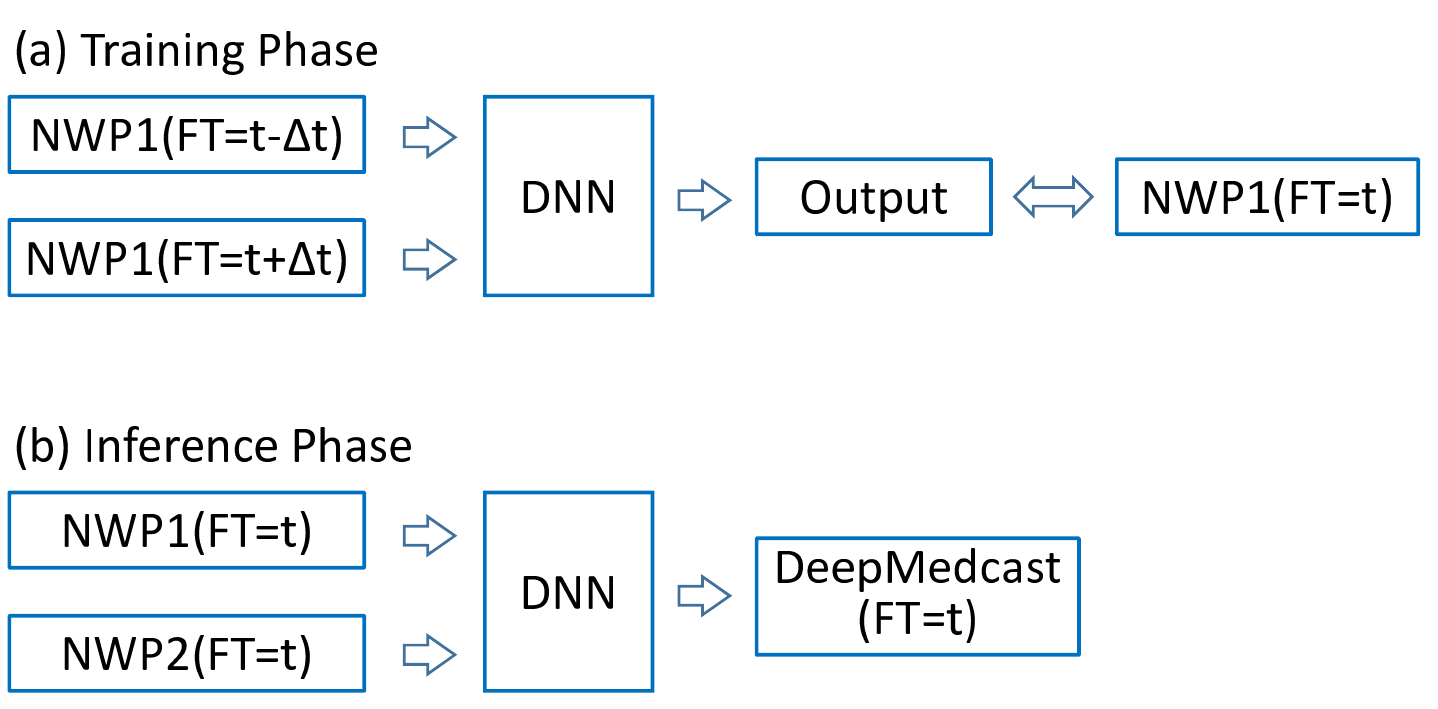}
  \caption{DeepMedcast framework for training and inference. (a) During the training phase, two forecast lead times from the same NWP model (NWP1 at FT~=~t~-~$\Delta$t and FT~=~t~+~$\Delta$t) are used as input, and the output from the DNN is compared with the same NWP model’s forecast at FT~=~t as the ground truth to train the network. (b) During the inference phase, predictions from two different NWP models (NWP1 and NWP2) at the same lead time (FT~=~t) are used as input to generate an intermediate forecast between the two models at FT~=~t.}
  \label{fig:fig1}
\end{figure}

\begin{figure}[t]
  \centering
  \includegraphics[width=0.70\textwidth,trim=0 0 0 0,clip]{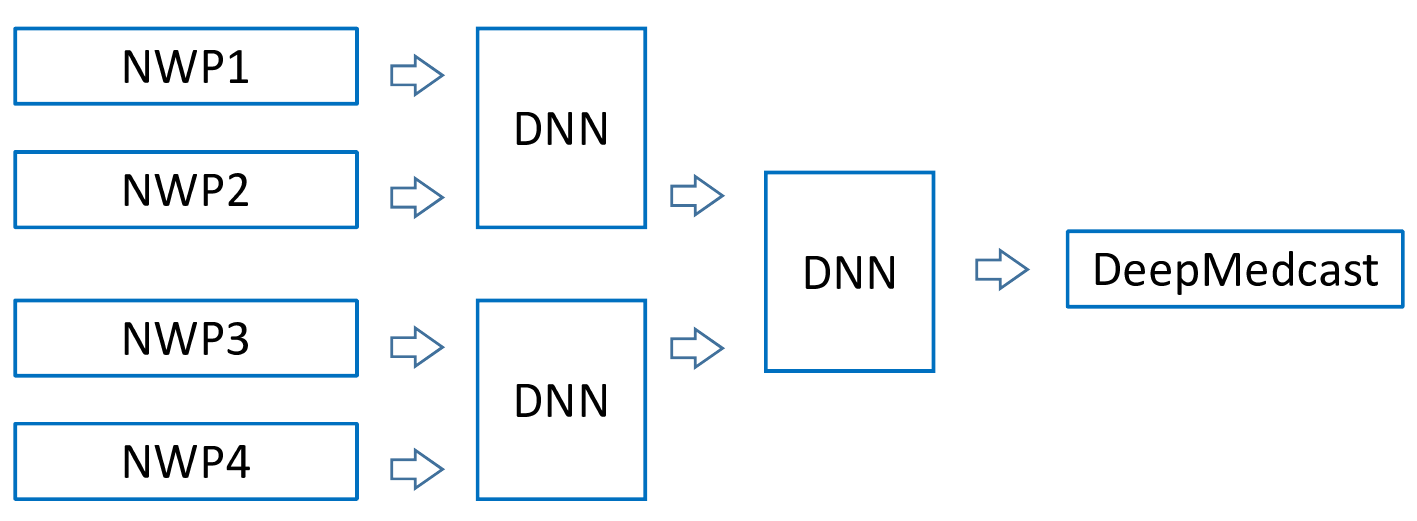}
  \caption{The recursive application of DeepMedcast, where intermediate forecasts are first generated between two NWP models (NWP1 and NWP2, NWP3 and NWP4), followed by the creation of an additional intermediate forecast between the outputs of the first two pairs.}
  \label{fig:fig2}
\end{figure}

\begin{figure}[t]
  \centering
  \includegraphics[width=0.65\textwidth,trim=0 0 0 0,clip]{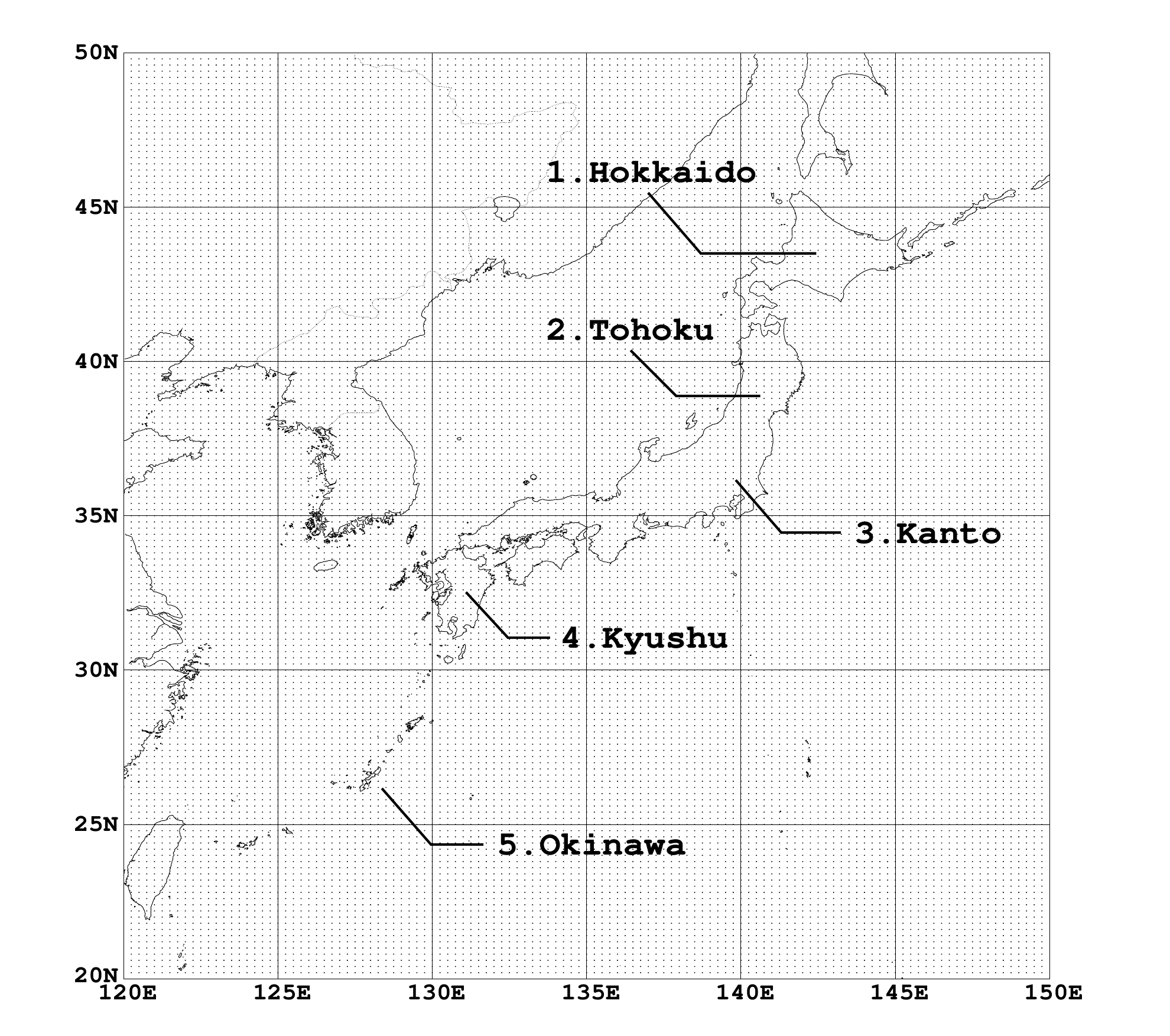}
  \caption{The target grid domain for this study. 121~$\times$~151 grids with 0.25-degree~$\times$~0.2-degree resolution around Japan. The dots on the map represent these grid points. The numbers and region names indicated in the figure are used in the case studies in Section 3.}
  \label{fig:fig3}
\end{figure}

\subsection{The framework of DeepMedcast}
\label{sec:sec2-1}
The main idea behind DeepMedcast lies in its original approach to generating intermediate forecasts between two NWP outputs. Figure 1 illustrates the framework of DeepMedcast. During the training phase, instead of using two different NWP outputs intended for creating an intermediate forecast, DeepMedcast utilizes data at two forecast lead times (FT), FT~=~t~-~$\Delta$t and FT~=~t~+~$\Delta$t, from a single NWP model as input variables for the deep neural network (DNN) (Fig. 1a). The output from the DNN is then compared to the forecast from the same NWP at the intermediate lead time (FT~=~t) to calculate the loss for the backpropagation process. This approach enables the network to generate intermediate forecasts while reducing the blurring effect often seen in machine learning (ML)-based post-processing, as the input variables are not inherently affected by errors relative to observed values, and the predictions at FT~=~t are expected to lie between those at FT~=~t ± $\Delta$t. During the inference phase, two different NWP outputs at the same forecast lead time are used to generate an intermediate forecast for the projection time (Fig. 1b). 

DeepMedcast is primarily designed to generate intermediate forecasts between two NWP models. However, the same DNN model can be applied recursively to generate intermediate forecasts between more than two NWP models. For instance, by taking intermediate forecasts between two pairs of NWP models, DeepMedcast can generate an intermediate result between four NWP models (Fig.2). This recursive approach could be extended further to create intermediate forecasts between 8, 16, or even more NWP models.

\begin{figure}[b]
  \centering
  \includegraphics[width=0.95\textwidth,trim=0 0 0 0,clip]{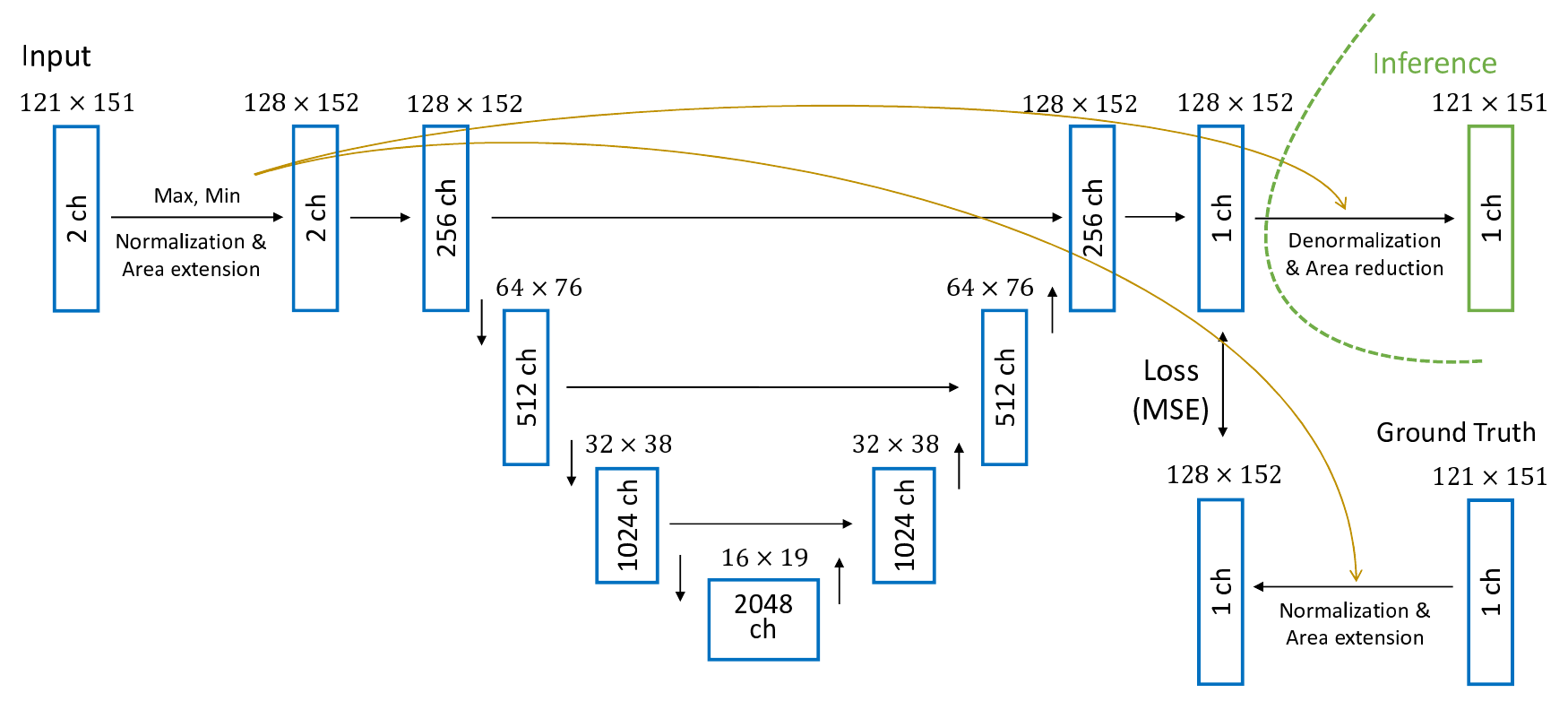}
  \caption{The DNN architecture used in DeepMedcast. The model takes two input channels and outputs a single channel. Input data is normalized using the maximum and minimum values, and during inference, the same values are applied for the denormalization process.}
  \label{fig:fig4}
\end{figure}

\subsection{Data used for the study}
\label{sec:sec2-2}
The NWP model used for training is GSM, which is operated by JMA four times a day (at 00, 06, 12, and 18 UTC as initial times). The training period spans nine years, from January 2013 to December 2021, while the validation period covers one year, from January to December 2022. GSM had a horizontal resolution of approximately 20~km until March 2023, after which it was upgraded to about 13~km \citep{JMA2024}. The GSM data used in this study is stored at JMA, where it is trimmed and linearly interpolated onto a 121~$\times$~151 grid with a resolution of 0.25 degrees by 0.2 degrees around Japan (Fig. 3). Hereafter, we refer to this as the target grid domain.

The forecast variables include wind components (U, V), temperature (T), and relative humidity (RH) at both the surface and the 700~hPa level, as well as mean sea-level pressure at the surface ($\mathrm{P}_\mathrm{sea}$). To reduce computational cost and execution time, each variable is used individually to train separate networks, with each network dedicated to a single variable. That means, each DNN model always takes two input channels (at FT~=~t~-~$\Delta$t and FT~=~t~+~$\Delta$t) and outputs one channel (at FT~=~t) for DeepMedcast. Both input channels are utilized by swapping their order, i.e., both FT~=~t~-~$\Delta$t and FT~=~t~+~$\Delta$t, as well as FT~=~t~+~$\Delta$t and FT~=~t~-~$\Delta$t, are employed to preserve symmetry. This strategy encourages the network to learn symmetric representations, so that meteorologically significant features—such as the center positions of TCs or extratropical cyclones, frontal boundaries, and shear lines—approximately align with the arithmetic mean of the corresponding features from the two input channels. The networks trained with 700~hPa data are employed to generate intermediate forecasts for the upper atmosphere (e.g., 850~hPa, 700~hPa, and 500~hPa) to reduce computational costs. The forecast lead times used in this study are t~=~9, 10, 11, 12, 13, and 14 hours, with $\Delta$t~=~±3 and ±6 hours, corresponding to t and $\Delta$t in Fig. 1, determined by taking both computational costs and accuracy into account.

For the case studies in Section 3, MSM, IFS, GraphCast, and Pangu-Weather are used along with GSM for inference. MSM is operated by JMA eight times a day (at 00, 03, ..., and 21 UTC as initial times) with a 5~km horizontal resolution, providing forecasts up to FT~=~78 hours for 00 and 12 UTC initial times and up to FT~=~39 hours for other initial times. IFS data, provided by ECMWF for the World Meteorological Organization (WMO) members, has a horizontal resolution of 0.5 degrees and is initialized four times daily at 00, 06, 12, and 18 UTC. Both GraphCast and Pangu-Weather have a horizontal resolution of 0.25 degrees, with data initialized at 00, 06, 12, and 18 UTC. These NWP outputs are linearly interpolated to the target grid domain for inference.

\subsection{DNN model architecture}
\label{sec:sec2-3}
In this study, a U-Net architecture \citep{Ronneberger2015} is applied as the DNN model. The structure of the network is illustrated in Fig. 4. The encoder part of the U-Net utilizes a convolutional network with kernel size~=~3, stride~=~1, and padding~=~1 for convolution operations. To progressively reduce the image size, MaxPooling layers with kernel size~=~2 and stride~=~2 are employed. This downsampling process continues until the image size is reduced to 1/8 of the original dimensions, at which point the channel count reaches 2048, starting from an initial 2 channels that are expanded to 256 channels. In each downsampling stage, the image size is halved while the number of channels doubles. In the decoder part, transposed convolutional layers with kernel size~=~2 and stride~=~2 are applied to upsample the feature maps, restoring the image to its original size while reducing the channel count by half at each stage. By the final stage, the image is returned to its original dimensions with 256 channels, which are then reduced to 1 channel in the output layer. The activation functions used in this network include rectified linear units (ReLU,\citep{Nair2010} for all layers except the output layer, which uses a sigmoid function to ensure output values are scaled between 0 and 1.

During the training phase, the two input channels and one ground truth channel, each consisting of 121~$\times$~151 grids, are normalized to a value range of 0 to 1 using the maximum and minimum values across all three channels. Specifically, for T, RH, and $\mathrm{P}_\mathrm{sea}$, the normalization is applied as:

\begin{equation}
x' = \frac{x-x_\mathrm{min}}{x_\mathrm{max}-x_\mathrm{min}}
\end{equation}

where $x'$ is the normalized value, $x$ is the input value, and $x_\mathrm{max}$ and $x_\mathrm{min}$ represent the maximum and minimum values, respectively. For wind components U and V, $x_\mathrm{max}$ and $x_\mathrm{min}$ are defined as:

\begin{eqnarray*}
x_\mathrm{max} &=& \mathrm{max}\left(|x_\mathrm{max}|, |x_\mathrm{min}|\right)  \\
x_\mathrm{min} &=&~-~x_\mathrm{max}
\end{eqnarray*}

and the same normalization is applied.

After normalization, the values are extended to 128~$\times$~158 grids by copying the last column and row to adjust to the network structure. The output values are compared with the normalized and extended ground truth values using the mean square error (MSE) as the loss function. We employ Adam \citep{Kingma2014} as optimization.

During the inference phase, the input values are normalized to the 0 to 1 range using the maximum and minimum values of the two input channels. The output values are then denormalized using the same maximum and minimum values, and resized back to 121~$\times$~151 grids by trimming the extended columns and rows, providing predictions at the target grid domain.

\begin{figure}[t]
  \centering
  \includegraphics[width=0.70\textwidth,trim=0 0 0 0,clip]{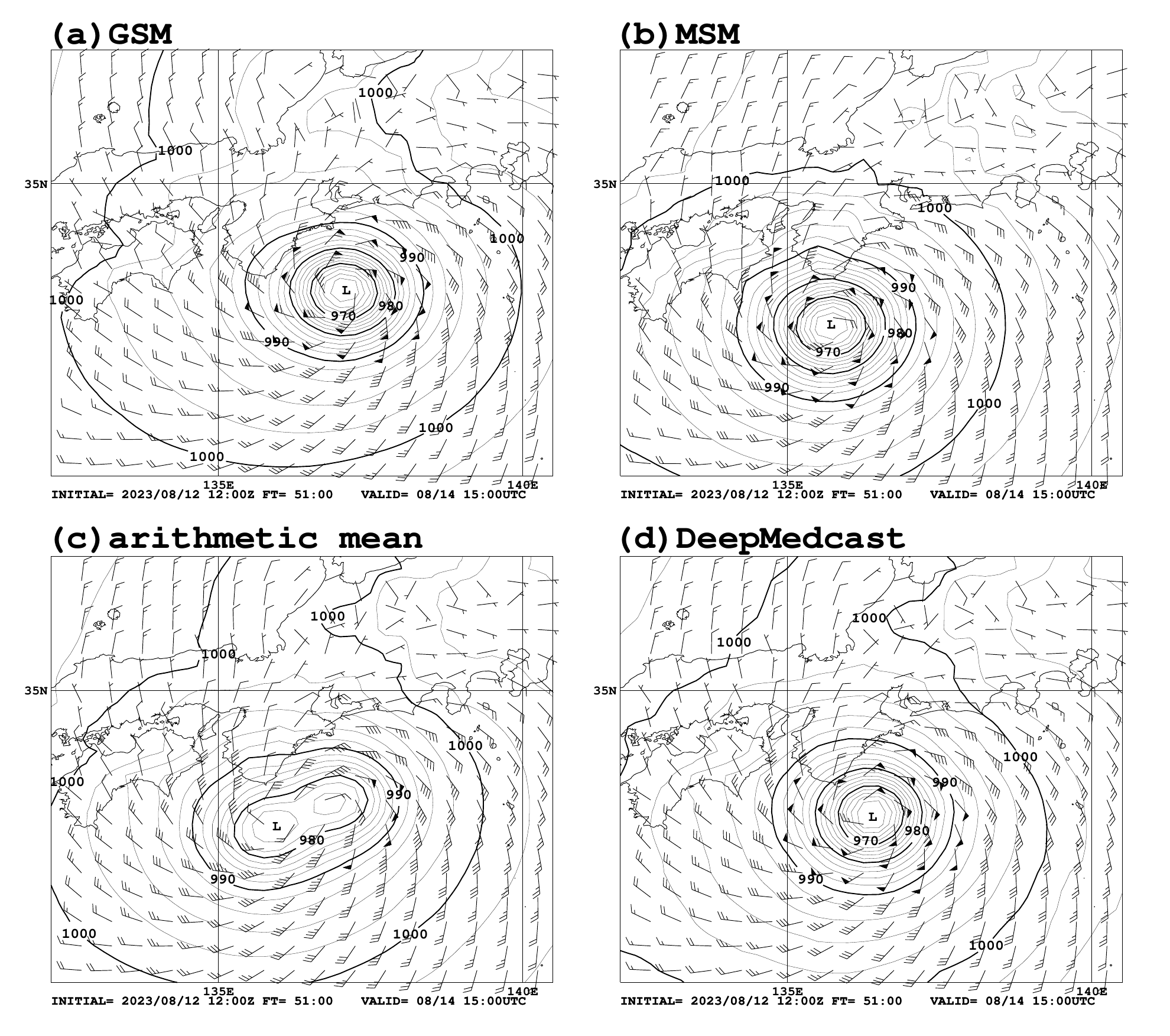}
  \caption{Comparison of Typhoon LAN predictions by (a) GSM, (b) MSM, (c) the arithmetic mean, and (d) DeepMedcast. The forecasts are based on the initial time of 12 UTC on 12 August 2023 with a forecast lead time of 51 hours. The black contours indicate mean sea-level pressure and wind barbs (units in kt) show surface winds.}
  \label{fig:fig5}

  \vspace{5mm}

  \centering
  \includegraphics[width=0.70\textwidth,trim=0 0 0 0,clip]{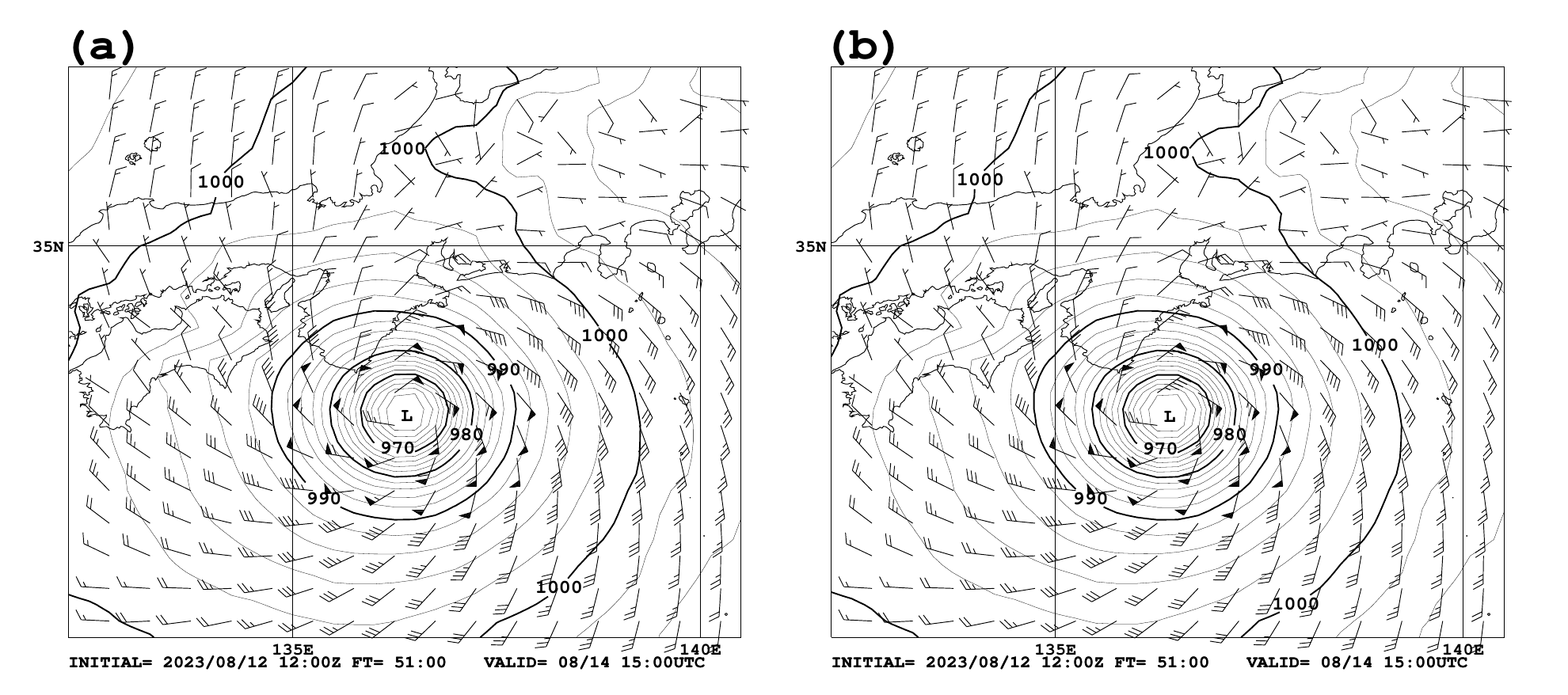}
  \caption{Comparison of DeepMedcast outputs with different input orders for the case in Fig. 5. (a) Result when GSM and MSM are provided in that order (same as Fig. 5d). (b) Result when the input order is reversed (MSM-GSM). While slight differences are present due to network asymmetry, the outputs remain qualitatively identical.}
  \label{fig:fig6}
\end{figure}

\begin{figure}[t]
  \centering
  \includegraphics[width=0.70\textwidth,trim=0 0 0 0,clip]{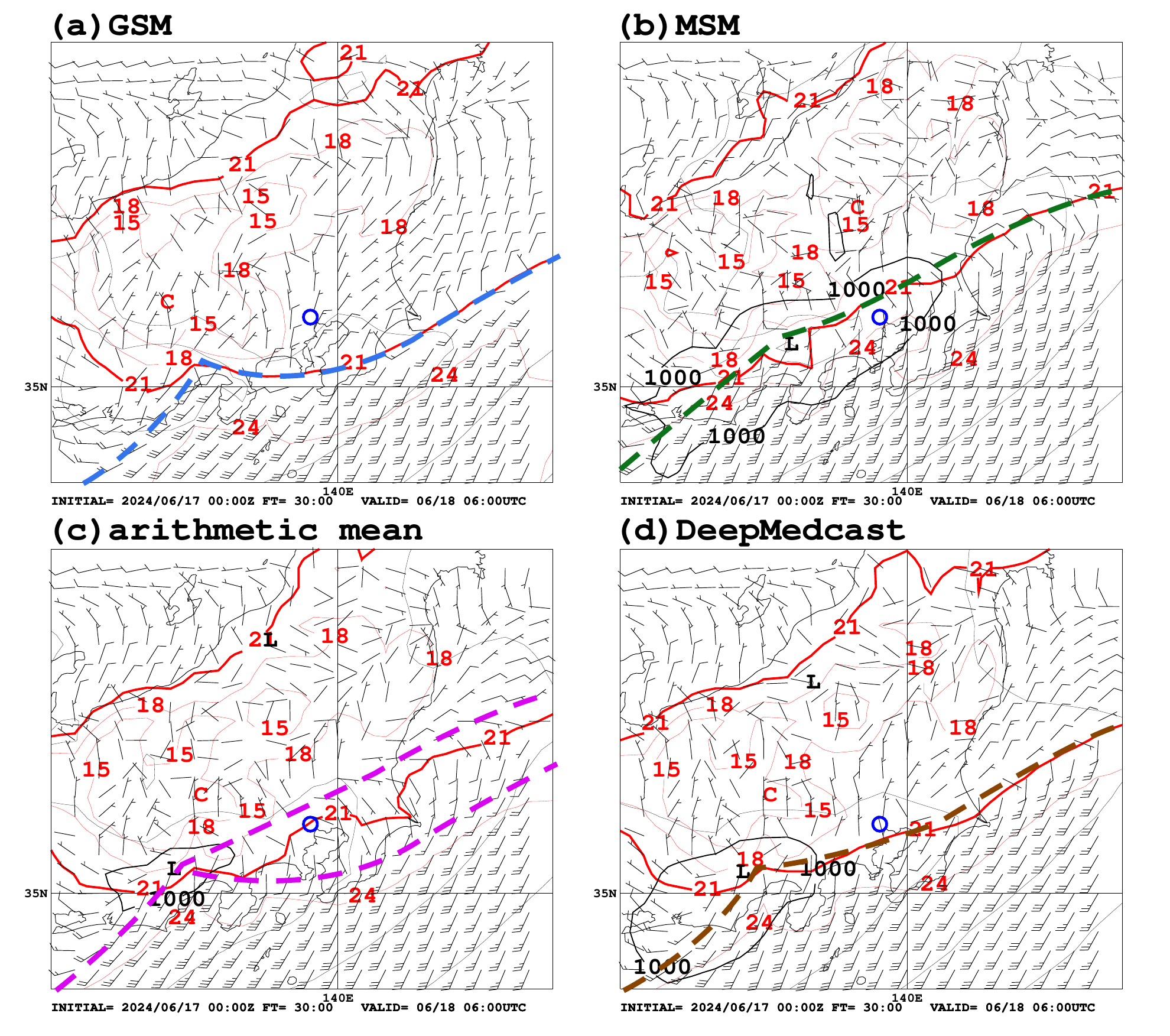}
  \caption{Comparison of predicted front positions by (a) GSM, (b) MSM, (c) the arithmetic mean, and (d) DeepMedcast. The forecasts are based on the initial time of 00 UTC on 17 June 2024 with a forecast lead time of 30 hours. The black contours indicate mean sea-level pressure, the red contours represent surface temperature, and wind barbs (units in kt) show surface winds. The blue, green, purple, and brown dashed lines represent the predicted front positions by GSM, MSM, the arithmetic mean, and DeepMedcast, respectively. Blue circles indicate the location of Tokyo.}
  \label{fig:fig7}
\end{figure}

\begin{figure}[htbp]
  \centering
  \includegraphics[width=0.70\textwidth,trim=0 0 0 0,clip]{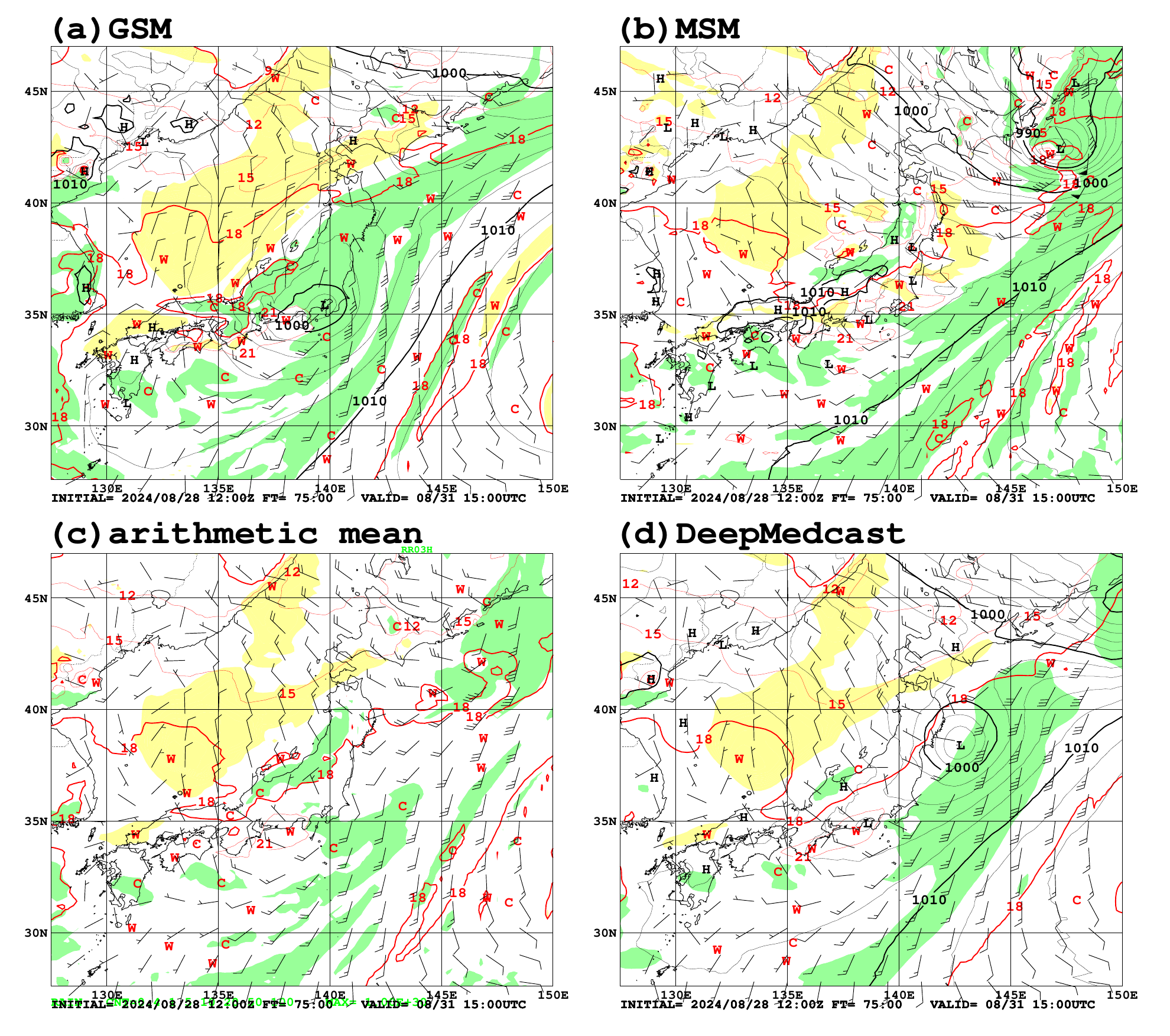}
  \caption{Comparison of predicted low-pressure systems by (a) GSM, (b) MSM, (c) the arithmetic mean, and (d) DeepMedcast. The forecasts are based on the initial time of 12 UTC on 28 August 2024 with a forecast lead time of 75 hours. The black contours indicate mean sea-level pressure, the wind barbs (units in kt) represent surface winds, the red contours show 850~hPa temperature, and the shaded regions in green and yellow highlight areas where the dew-point depression at 850~hPa is below 3$^\circ$C and above 15$^\circ$C, respectively.}
  \label{fig:fig8}
\end{figure}

\begin{figure}[t]
  \centering
  \includegraphics[width=1.0\textwidth,trim=0 0 0 0,clip]{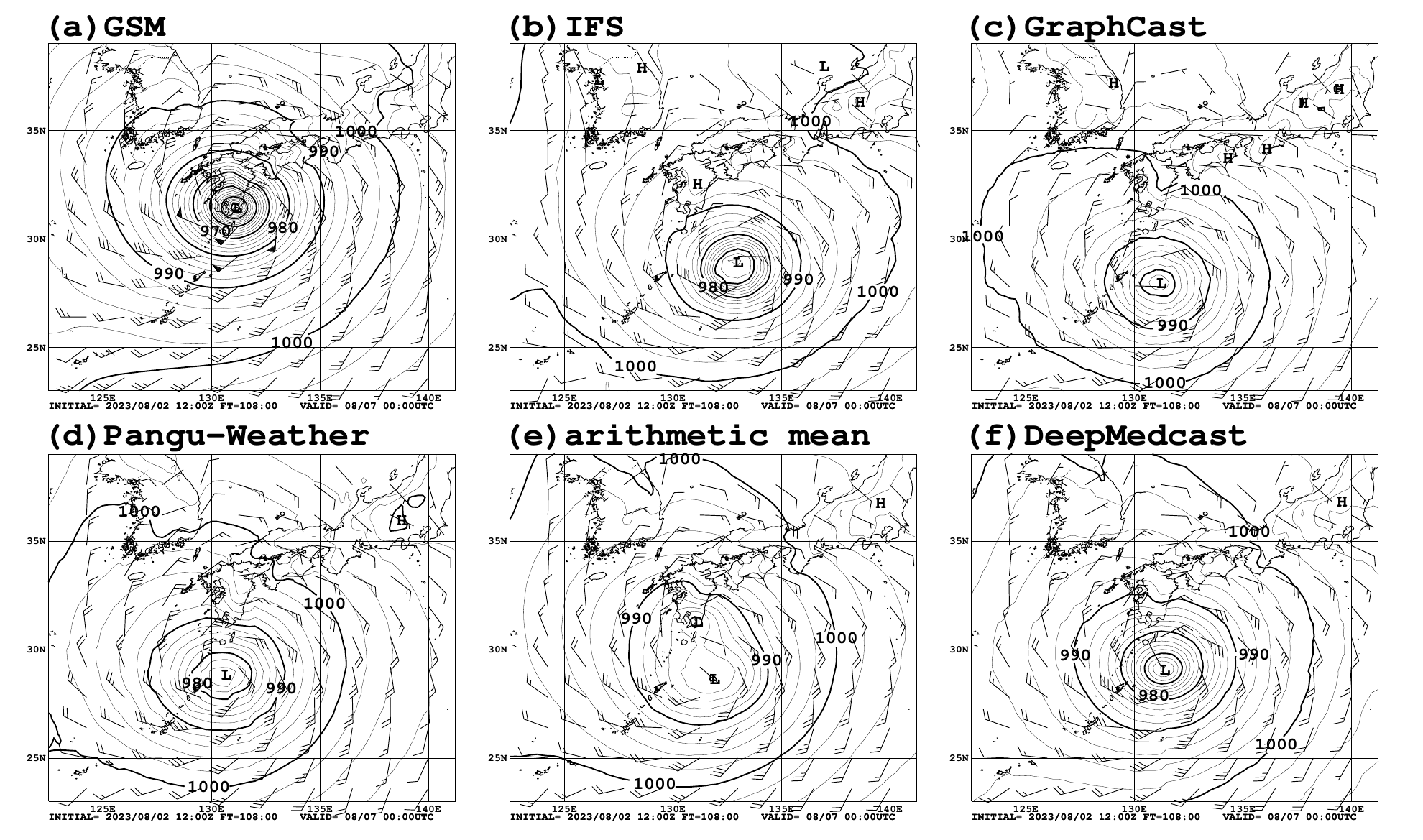}
  \caption{Comparison of Typhoon KHANUM predictions by (a) GSM, (b) IFS, (c) GraphCast, (d) Pangu-Weather, (e) the arithmetic mean, and (f) DeepMedcast. The predictions are based on the initial time of 12 UTC on 2 August 2023 with a forecast lead time of 108 hours. The black contours indicate mean sea-level pressure and wind barbs (units in kt) show surface winds.}
  \label{fig:fig9}

  \vspace{5mm}

  \centering
  \includegraphics[width=1.0\textwidth,trim=0 0 0 0,clip]{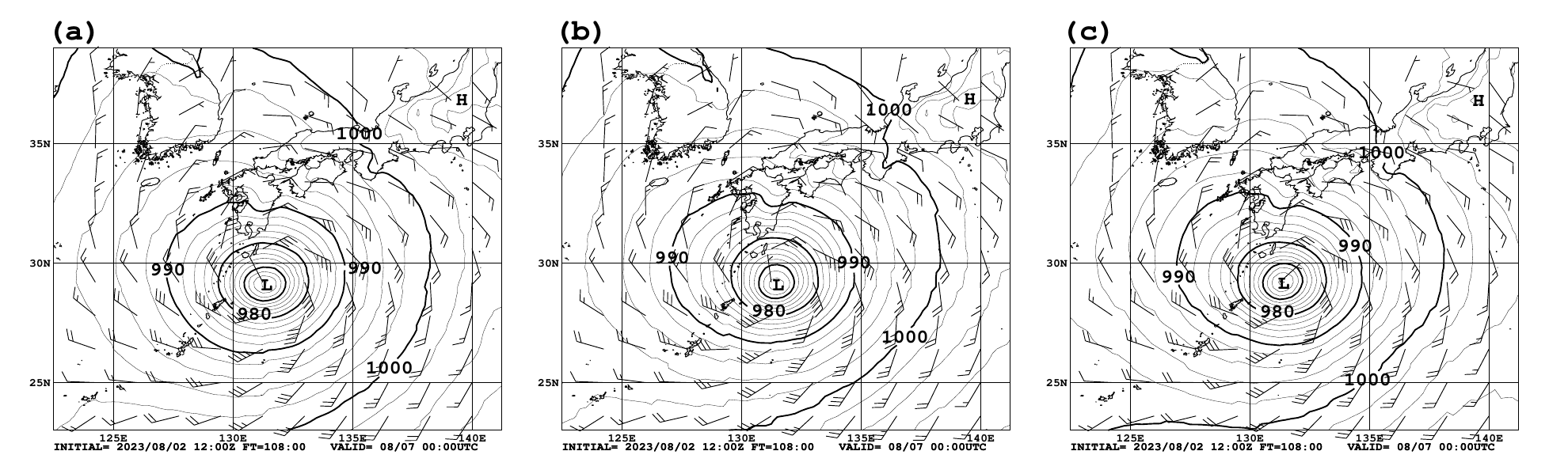}
  \caption{DeepMedcast forecasts based on the initial time of 12 UTC on 2 August 2023 with a forecast lead time of 108 hours for Typhoon KHANUN using different orders of intermediate forecast generation from the four NWP models: GSM, IFS, GraphCast, and Pangu-Weather. (a) Intermediate forecasts are first taken between GSM and IFS, and between GraphCast and Pangu-Weather, then combined. (b) GSM–GraphCast and IFS–Pangu-Weather. (c) GSM–Pangu-Weather and IFS–GraphCast.}
  \label{fig:fig10}
\end{figure}

\section{Results}
\label{sec:sec3}
In this section, we demonstrate the capability of DeepMedcast through four case studies and verification results. The forecast data used here is from a period beginning in January 2023, which is independent of the DNN’s training and validation periods. The case studies compare the atmospheric fields generated by DeepMedcast with those obtained via arithmetic mean of the NWP outputs.

\subsection{Case 1: Position discrepancy in a typhoon forecast between GSM and MSM}
\label{sec:sec3-1}
The first case study examines a typhoon forecast where there is a positional discrepancy between GSM and MSM. Figure 5 shows the predictions at FT~=~51 hours based on the initial time of 12 UTC on 12 August 2023. This case focuses on Typhoon LAN which was moving northwest over the ocean south of Japan. At FT~=~51 hours, GSM predicted the typhoon’s position at 33.3$^\circ$N, 137.1$^\circ$E, while MSM placed it southwest at 32.7$^\circ$N, 135.7$^\circ$E. Both models predicted a central pressure of 960~hPa, with the maximum wind speed of 79~kt (1~kt~$\simeq$~0.514~m s$^{-1}$) (GSM) and 68~kt (MSM) (Figs. 5a and 5b).

When the mean sea-level pressure and surface wind components from GSM and MSM were averaged arithmetically, the typhoon’s center split into two, aligning with the predicted positions from each model (Fig. 5c). Such a result is evidently unrealistic and lacks interpretability. The central pressure weakened to 974~hPa, and the maximum wind speed reduced to 58~kt, which made the forecast meteorologically unnatural, with the typhoon taking on an elongated structure and weakening wind speeds near the center. This resulted in a forecast that was difficult to explain and potentially misleading.

In contrast, DeepMedcast generated a plausible forecast, placing the typhoon at 33.0$^\circ$N, 136.4$^\circ$E, halfway between GSM and MSM predictions (Fig. 5d). The typhoon maintained a single, natural, and interpretable shape with a central pressure of 960~hPa and the maximum wind speed of 69~kt, representing an intermediate forecast between the two NWP models.

Figure 6 compares the DeepMedcast outputs with different input orders. Figure 6a shows the result when GSM and MSM are used as inputs in that order (identical to Fig. 5d), while Fig. 6b shows the result when the input order is reversed (MSM-GSM). As expected, the predicted structure—including the typhoon center position, central pressure, and surrounding wind field—remains qualitatively consistent, despite minor differences due to the asymmetry of the trained neural network. This indicates that while DeepMedcast is not strictly order-invariant, the resulting intermediate forecasts are robust to changes in input order.

\subsection{Case 2: Discrepancy in a front position forecast between GSM and MSM}
\label{sec:sec3-2}
The second case study examines a forecast where there was a positional discrepancy in the predicted location of a front between GSM and MSM. Figure 7 shows the predictions at FT~=~30 hours based on the initial time of 00 UTC on 17 June 2024. At the initial time, a stationary front was located south of Japan (not shown), and by FT~=~30 hours, the front was predicted to move northward toward Tokyo (indicated by the blue circles in the figure).

GSM predicted the front to the south of Tokyo, indicated by the blue dashed line, with a clear wind direction and speed shear, which corresponds well with the 21$^\circ$C isotherm around Tokyo (Fig. 7a). In contrast, MSM placed the front north of Tokyo (green dashed line), also with a clear wind direction and speed shear aligned with the 21$^\circ$C isotherm (Fig. 7b), resulting in a positional discrepancy between the two NWP models. Consequently, GSM predicted a northerly to northeasterly wind and cooler temperatures around Tokyo, while MSM predicted southerly to southwesterly winds and warmer temperatures, leading to significant differences in the forecast for Tokyo.

The arithmetic mean of the GSM and MSM predictions (Fig. 7c) results in a split structure for the front, with wind shear corresponding to the locations predicted by GSM and MSM (shown by the purple lines), while the 21$^\circ$C isotherm is predicted between the two fronts. This demonstrates that when there is a discrepancy in the predicted front position, simple averaging of the atmospheric fields leads to an unnatural and uninterpretable forecast that cannot maintain the original front structure.

In contrast, DeepMedcast (Fig. 7d) generates a clear wind direction and speed shear at the intermediate position between the GSM and MSM predictions (indicated by the brown dashed line), which corresponds well with the 21$^\circ$C isotherm. DeepMedcast successfully produces a realistic and interpretable intermediate forecast while preserving the structure of the original front.

\subsection{Case 3: Significant difference in low-pressure system position between GSM and MSM}
\label{sec:sec3-3}
The third case study highlights a situation where there was a large difference in the predicted position of a low-pressure system between GSM and MSM. Figure 8 shows surface wind and mean sea-level pressure, along with temperature and dew-point depression at 850~hPa, at FT~=~75 hours based on the initial time of 12 UTC on 28 August 2024. At the initial time, Typhoon SHANSHAN was located south of Kyushu (see Fig. 3, location 4) at 30.6$^\circ$N, 130.2$^\circ$E, slowly moving northward (not shown). By 15 UTC on 31 August (FT~=~75 hours), the system, which had either remained a tropical storm or transitioned into a low-pressure system, was predicted by GSM to be south of the Kanto region (see Fig. 3, location 3) at 35.4$^\circ$N, 139.8$^\circ$E (Fig. 8a), while MSM placed it east of Hokkaido (see Fig. 3, location 1) at 42.4$^\circ$N, 147.3$^\circ$E (Fig. 8b).

This case highlights a large positional difference of about 1000~km between the GSM and MSM predictions. When the arithmetic mean of these is taken (Fig. 8c), it results in two distinct low-pressure systems at the positions predicted by each model, creating an uninterpretable forecast. In contrast, DeepMedcast predicted a single low-pressure system located between the two forecasts, at 39.5$^\circ$N, 143.0$^\circ$E, off the Pacific coast of Tohoku (see Fig. 3, location 2; Fig. 8d). Additionally, when examining the moisture area at 850~hPa (dew-point depression < 3$^\circ$C), the arithmetic mean shows moist areas surrounding both the GSM and MSM low-pressure systems, with a relatively dry region in between. On the other hand, DeepMedcast represents a moist area around its low-pressure system, corresponding well with the surface pressure field, providing a realistic and interpretable forecast.

\subsection{Case 4: Intermediate forecast between four NWP models for typhoon KHANUN}
\label{sec:sec3-4}

\begin{table}
  \centering
  \caption{Comparison of Typhoon KHANUN predictions from four NWP models, their arithmetic mean, and DeepMedcast. The table presents the predicted central position, central pressure, and maximum wind speed based on the initial time of 12 UTC on 2 August 2023 with a forecast lead time of 108 hours.}
  \label{tab:tab1}
  \hspace{-1mm}
  \begin{tabular}{|c|c|c|c|c|}
    \hline\rule[0mm]{0mm}{3.75mm}
    & Model & Central position & Central pressure & Maximum wind speed \\
    \hline\rule[0mm]{0mm}{3.75mm}
    (a) & GSM & 31.4$^\circ$N, 131.1$^\circ$E & 938~hPa & 68~kt \\
    \hline\rule[0mm]{0mm}{3.75mm}
    (b) & IFS & 28.9$^\circ$N, 133.0$^\circ$E & 966~hPa & 51~kt \\
    \hline\rule[0mm]{0mm}{3.75mm}
    (c) & GraphCast & 28.0$^\circ$N, 131.2$^\circ$E & 977~hPa & 36~kt \\
    \hline\rule[0mm]{0mm}{3.75mm}
    (d) & Pangu-Weather & 28.8$^\circ$N, 130.7$^\circ$E & 975~hPa & 39~kt \\
    \hline\rule[0mm]{0mm}{3.75mm}
    (e) & arithmetic mean & --- & 979~hPa & 36~kt \\
    \hline\rule[0mm]{0mm}{3.75mm}
    (f) & DeepMedcast & 29.3$^\circ$N, 131.4$^\circ$E & 964~hPa & 42~kt \\
    \hline
  \end{tabular}
\end{table}

The fourth case study presents an intermediate forecast between four NWP models: GSM, IFS, GraphCast, and Pangu-Weather. Figure 9 shows surface wind and mean sea-level pressure at FT~=~108 hours, based on the initial time of 12 UTC on 2 August 2023. At the initial time, Typhoon KHANUN was located west of Okinawa (see Fig. 3, location 5) at 26.2$^\circ$N, 125.6$^\circ$E, slowly moving westward (not shown). By 00 UTC on 7 August (FT~=~108 hours), the central position was predicted by the four models to be at 31.4$^\circ$N, 131.1$^\circ$E (GSM), 28.9$^\circ$N, 133.0$^\circ$E (IFS), 28.0$^\circ$N, 131.2$^\circ$E (GraphCast), and 28.8$^\circ$N, 130.7$^\circ$E (Pangu-Weather). The typhoon central position, central pressure, and maximum wind speed at FT~=~108 hours for each model are summarized in Table 1.

When the arithmetic mean of the four models is taken, the center splits into two, with a weakened central pressure of 979~hPa and the maximum wind speed of 36~kt (Fig. 9e), both the same or weaker than the predictions of any individual model. In contrast, DeepMedcast predicted a single center at 29.3$^\circ$N, 131.4$^\circ$E, with the central pressure of 964~hPa and the maximum wind speed of 42~kt (Fig. 9f), representing an intermediate intensity forecast between the four NWP models. The average central latitude, longitude, and pressure of the four NWP models were 29.3$^\circ$N, 131.5$^\circ$E, and 964~hPa, respectively, matching DeepMedcast's prediction. The average maximum wind speed of the four models was 49~kt, meaning DeepMedcast’s forecast was slightly weaker than the average.

Figure 10 illustrates the effect of changing the order in which intermediate forecasts are taken when combining the four NWP models. Figure 10a is identical to Fig. 9f and shows the result when intermediate forecasts are first generated between GSM and IFS, and between GraphCast and Pangu-Weather, followed by taking an intermediate forecast between those two results. Figure 10b shows the case where the intermediate forecasts are first taken between GSM and GraphCast, and between IFS and Pangu-Weather, then combined. Figure 10c presents the result when intermediate forecasts are first taken between GSM and Pangu-Weather, and between IFS and GraphCast. As in the two-model case discussed in Section 3.1, the outputs differ slightly due to the inherent asymmetry of the trained network and the recursive nature of the procedure. However, the predicted typhoon structure, including its central position, pressure, and wind field, remains qualitatively consistent across all three cases. This suggests that although DeepMedcast is not strictly order-invariant, it produces robust intermediate forecasts in practice.

\subsection{Statistical evaluation of DeepMedcast using surface wind observations}
This section presents the verification results of DeepMedcast generated from GSM and MSM. Surface wind predictions from DeepMedcast, along with the input GSM and MSM forecasts, were verified against observations from the Automated Meteorological Data Acquisition System (AMeDAS), an automated observation network operated by JMA. The verification metric is the root mean square error (RMSE), defined as follows:

\begin{equation}
\mathrm{RMSE} = \sqrt{\frac{1}{T}\sum^T_{t=1}\frac{1}{N}\sum^N_{n=1}\left( F_{nt}~-~O_{nt} \right)^2}
\end{equation}

where $T$ nad $N$ are the numbers of forecast times and stations used for verification, and $F_{nt}$ and $O_{nt}$ represent the forecast and observed winds at station $n$ and time $t$, respectively. Predictions from DeepMedcast, GSM, and MSM were linearly interpolated to each AMeDAS station from the four surrounding grid points.

Figure 11 shows the RMSE of wind speed (Fig. 11a) and wind direction (Fig. 11b) by forecast lead time, ranging from 3 to 39 hours. The verification was conducted over one year, from January to December 2023, using forecasts initialized four times daily (00, 06, 12, and 18 UTC), independent of the training and validation periods. In both panels, the red, blue, and green lines represent DeepMedcast, GSM, and MSM, respectively. As shown in Fig. 11, DeepMedcast achieves lower RMSEs for both wind speed and direction across all forecast lead times compared to its input models. The RMSE for wind direction in Fig. 11b shows a 6-hourly fluctuation pattern, reflecting the four-times-daily initialization and diurnal variation.

\begin{figure}[t]
  \centering
  \includegraphics[width=0.80\textwidth,trim=0 0 0 0,clip]{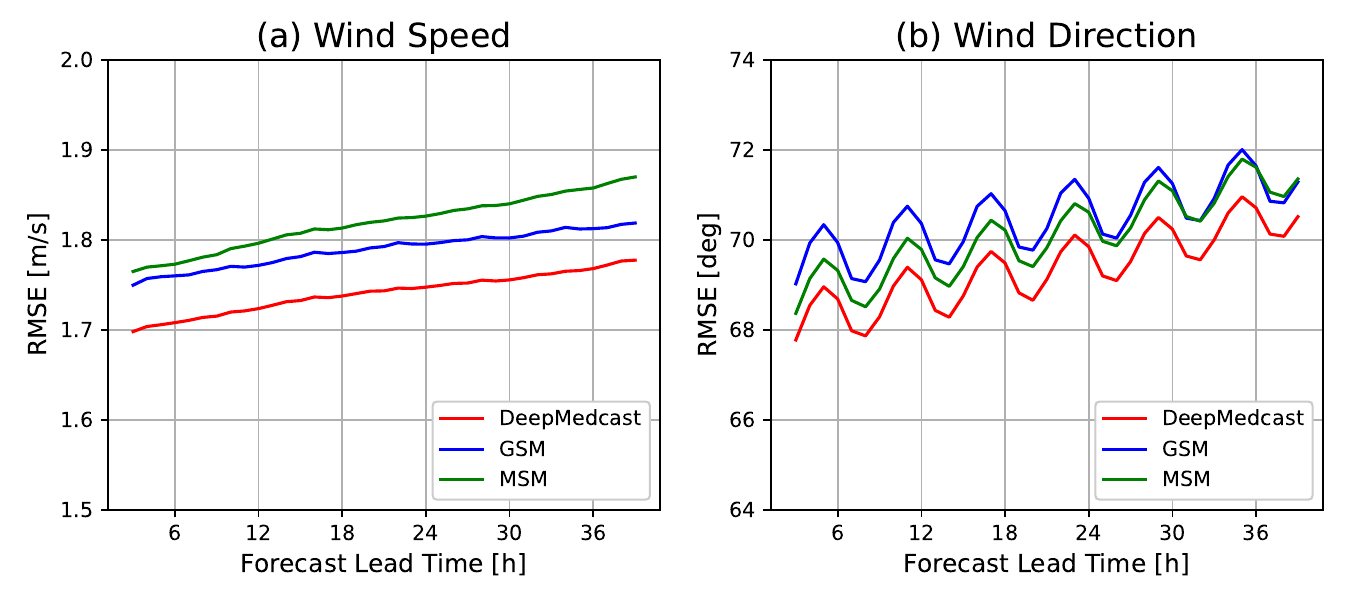}
  \caption{Root mean square error (RMSE) of (a) surface wind speed and (b) surface wind direction for DeepMedcast (red), input GSM (blue), and input MSM (green) forecasts, verified against AMeDAS observations. The verification period spans one year, from January to December 2023, with forecasts initialized at 00, 06, 12, and 18 UTC.}
  \label{fig:fig11}
\end{figure}

\section{Discussion}
\label{sec:sec4}
\subsection{Contributions to operational forecasting}
\label{sec:sec4-1}
As demonstrated by the case studies and verification in Section 3, DeepMedcast is capable of generating plausible and interpretable intermediate forecasts. The ability to generate intermediate forecasts between multiple models is expected to significantly contribute to operational forecasting. As mentioned in the Introduction, TC track forecasts, which are based on consensus from multiple NWP models, specifically the average position of the TC center predicted by these models, serve as primary reference in JMA's operational forecasting. Consequently, forecasters responsible for general, marine, and aviation forecasts must ensure that their forecasts align with the TC track forecasts. However, since no NWP model inherently conforms to the TC track forecasts, forecasters must adjust the existing NWP outputs in their minds to construct forecast scenarios that follow the TC track forecasts. DeepMedcast has a capability to provide two-dimensional wind and pressure fields that align with TC track forecasts, which could greatly improve the efficiency and standardization of tasks for operational forecasting.

Additionally, DeepMedcast can be effectively utilized in operational forecasting when there are discrepancies in the predicted positions of low-pressure systems or fronts among multiple models. When significant differences exist among NWP models, forecasters need to choose between two options: either using one model as the main scenario while treating others as alternative scenarios, or applying averaging methods. By generating an intermediate state between two or more NWP models, DeepMedcast provides a forecast scenario that is more plausible than individual NWP models. However, it should be noted that it does not explicitly represent the variability among the original NWP models. This issue is not unique to DeepMedcast but is also present when using arithmetic or weighted averaging of multiple NWP models or post-processing techniques. Since variability is an indicator of forecast uncertainty, especially for longer lead times, it is important to take this information into account in daily operational forecasting. One practical approach is to incorporate the values of the original NWP models and their spread alongside the intermediate forecast. This framework can further enhance operational forecasting by offering a practical way to take model variability into account.

\subsection{Key Features of DeepMedcast’s Architecture}
\label{sec:sec4-2}
There are two important features associated with DeepMedcast's architecture. The first is its flexibility in increasing the amount of training data. As mentioned in Section 2, this study used t~=~9–14 and $\Delta$t~=~±3, ±6, and in our experience, increasing t and $\Delta$t leads to better forecast representation. One common issue in training DNN models is a lack of sufficient training data (e.g., \cite{Deng2009, LeCun2015}). However, in the case of DeepMedcast, more training data can easily be generated by increasing t and $\Delta$t or by adding additional NWP models. It is important to note, though, that increasing t and $\Delta$t requires more memory and computational time, which should be considered when expanding the dataset.

The second is DeepMedcast’s maintainability. Despite being trained solely on 20-km resolution GSM data, DeepMedcast works effectively not only with 13-km resolution GSM data but also with other NWP models such as MSM, IFS, GraphCast, and Pangu-Weather. This is significant because most AI or ML methods in meteorology learn the relationship between input and target data, and when the characteristics of the input data change due to NWP model updates, retraining, fine-tuning, and/or online learning are usually required. While this is an unavoidable task for most AI or ML methods in meteorology, it is a time-consuming yet essential task that operational centers have traditionally managed. However, DeepMedcast can be applied to various NWP models without updating the DNN model since it is not designed to correct NWP model biases, which significantly reduces the maintenance costs for operational centers.

\section{Summary}
\label{sec:sec5}
In this paper, we introduced DeepMedcast, a novel deep learning-based approach for producing intermediate forecasts between two or more NWP models. DeepMedcast was developed to generate plausible and interpretable intermediate forecast, bridging the gap between NWP model outputs.

A key advantage of DeepMedcast is its applicability to various NWP outputs without the need for retraining or fine-tuning the DNN. By providing plausible intermediate forecasts, DeepMedcast can significantly enhance the efficiency and standardization of operational forecasting tasks, including general, marine, and aviation forecasts.

Although DeepMedcast introduces some advancements, further research and development are needed to address several challenges. In this study, U-Net was employed as the DNN architecture; however, advanced methods such as Transformers \citep{Vaswani2017, Dosovitskiy2020} and Diffusion models \citep{Song2019, Ho2020} may further enhance DeepMedcast's representational capabilities. As shown in the case study in Section 3.4, the current method tends to slightly underestimate the maximum wind speed near TCs. Enhancing the DNN could help resolve this issue. Additionally, while this study trained separate networks for each physical variable to reduce computational cost, incorporating multiple physical variables as input could potentially enhance forecast accuracy. By developing post-processing methods that use DeepMedcast as input, it would be possible to provide even more accurate predictions. This study focused on generating intermediate forecasts using a 1:1 weighting ratio, meaning that the two input models were given equal weight. Future work should explore methods for generating intermediate forecasts with other weighting ratios, such as 1:2. This would enable the application of DeepMedcast to cases involving several models that is not a power of two, such as finding an intermediate forecast among three models. Furthermore, while this study demonstrated intermediate forecasts using two or four NWP models, DeepMedcast could be extended to 8, 16, or more inputs, enabling the use of multiple NWP and ensemble models. Lastly, this study did not address intermediate precipitation forecasts. Since precipitation is one of the most critical variables in weather forecasting, future work will focus on developing intermediate precipitation forecasts.

\section*{Data Availability Statement}
The datasets used in this study are available from the following sources: The GSM and MSM data are operationally produced by JMA and can be accessed through the Japan Meteorological Business Support Center (\url{http://www.jmbsc.or.jp/en/index-e.html}). The IFS data are accessible to WMO members via the ECMWF website (\url{https://www.ecmwf.int/en/forecasts/datasets/wmo-additional}). The GraphCast and Pangu-Weather source code and plugins are available under open-source licenses in the ECMWF GitHub repository (\url{https://github.com/ecmwf-lab/ai-models}). Pre-trained models of Pangu-Weather and GraphCast, used without modification to generate forecast data, are specifically accessible at \url{https://github.com/ecmwf-lab/ai-models-panguweather} and \url{https://github.com/ecmwf-lab/ai-models-graphcast}.

\section*{Acknowledgements}
We acknowledge ECMWF for providing IFS data available to WMO members through their website (\url{https://www.ecmwf.int/en/forecasts/datasets/wmo-additional}). Additionally, we are grateful for the availability of GraphCast and Pangu-Weather and extend our thanks to their respective developers. GraphCast and Pangu-Weather were used without modification to generate forecast data and are accessible via the ECMWF AI GitHub repository (\url{https://github.com/ecmwf-lab/ai-models}), supported by ECMWF. The author declares no conflicts of interest associated with this manuscript.

\bibliographystyle{unsrtnat}


\begin{thebibliography}{38}
\providecommand{\natexlab}[1]{#1}
\providecommand{\url}[1]{\texttt{#1}}
\expandafter\ifx\csname urlstyle\endcsname\relax
  \providecommand{\doi}[1]{doi: #1}\else
  \providecommand{\doi}{doi: \begingroup \urlstyle{rm}\Url}\fi

\bibitem[Agueh and Carlier(2011)]{Agueh2011}
Agueh, M. and G. Carlier, 2011: Barycenters in the Wasserstein space SIAM Journal on Mathematical Analysis, \textbf{43(2)}, 904–924. [Available at \url{https://hal.science/hal-00637399v1/document}.]

\bibitem[Bi et al.(2022)]{Bi2022}
Bi,K.,L. Xie,H.Zhang, X.Chen, X.Gu, and Q.Tian, 2022: Pangu-Weather: A 3D High-Resolution System for Fast and Accurate Global Weather Forecast. \textit{arXiv preprint arXv:2211.02556.}

\bibitem[Bi et al.(2023)]{Bi2023}
Bi, K., L. Xie, H. Zhang, X. Chen, X. Gu, and Q. Tina, 2023: Accurate medium-range global weather forecasting with 3D neural networks. \textit{Nature}, \textbf{619} 533–538. \doi{10.1038/s41586-023-06185-3}.

\bibitem[Bodnar et al.(2024)]{Bodnar2024}
Bodnar, C., W. P. Bruinsma, A. Lucic, M. Stanley, J. Brandstetter, P. Garvan, M. Riechert, J. Weyn, H. Dong, A. Vaughan, J. K. Gupta, K. Tambiratnam, A. Archibald, E. Heider, M. Welling, R. E. Turner, P. Perdikaris, 2024: Aurora: A Foundation Model of the Atmosphere. \textit{arXiv preprint arXiv:2405.13063.}

\bibitem[Bonev et al.(2023)]{Bonev2023}
Bonev, B., T. Kurth, C. Hundt, J. Pathak, M. Baust, K. Kashinath, A. Anandkumar, 2023: Spherical Fourier Neural Operators: Learning Stable Dynamics on the Sphere. \textit{arXiv preprint arXiv:2306.03838.}

\bibitem[Brown et al.(2012)]{Brown2012}
Brown, A., S. Milton, M. Cullen, B. M. J. Golding, and A. Shelly, 2012: Unified modeling and prediction of weather and climate: a 25 year journey. \textit{Bull. Amer. Meteor. Soc.}, \textbf{93}, 1865–1877.

\bibitem[Cangialosi et al.(2023)]{Cangialosi2023}
Cangialosi, J., B.J. Reinhart, and J. Martinez, 2023: \textit{National Hurricane Center verification report, 2023 Hurricane Season.} Natinal Hurricane Center, 81pp. [Available at \url{https://www.nhc.noaa.gov/verification/pdfs/Verification\_2023.pdf}.]

\bibitem[Chen et al.(2023)]{Chen2023}
Chen, K., T. Han, J. Gong, L. Bai, F. Ling, J. Luo, X. Chen, L. Ma, T. Zhang, R. Su, Y. Ci, B. Li, X. Yang, and W. Ouyang, 2023: FengWu: Pushing the Skillful Global Medium-range Weather Forecast beyond 10 Days Lead. \textit{arXiv preprint arXiv:2304.02948.}

\bibitem[Deng et al.(2009)]{Deng2009}
Deng, J., W. Dong, R. Socher, L. Li, K. Li, and L. Fei-Fei, 2009: ImageNet: A large-scale hierarchical image database. IEEE Conference on Computer Vision and Pattern Recognition, 248–255.

\bibitem[Dosovitskiy et al.(2020)]{Dosovitskiy2020}
Dosovitskiy, A., L. Beyer, A. Kolesnikov, D. Weissenborn, X. Zhai, T. Unterthiner, M. Dehghani, M. Minderer, G. Heigold, S. Gelly, J. Uszkoreit, and N. Houlsby, 2020: An Image is Worth 16x16 Words: Transformers for Image Recognition at Scale. \textit{arXiv preprint arXiv:2010.11929.}

\bibitem[Dowell et al.(2022)]{Dowell2022}
Dowell, D. C., C. R. Alexander, E. P. James, S. S. Weygandt, S. G. Benjamin, G. S. Manikin, B. T. Blake, J. M. Brown, J. B. Olson, M. Hu, T. G. Smirnova, T. Ladwig, J. S. Kenyon, R. Ahmadov, D. D. Turner, J. D. Duda, T. I. Alcott, 2022: The High-Resolution Rapid Refresh (HRRR): An hourly updating convection-allowing forecast model. Part I: Motivation and system description. \textit{Wea. Forecasting}, \textbf{37}, 1371–1395, \doi{10.1175/WAF-D-21-0151.1}.

\bibitem[Duc and Sawada(2024)]{Duc2024}
Duc, L. and Y. Sawada, 2024: Geometry of rainfall ensemble means: from arithmetic averages to Gaussian-Hellinger barycenters in unbalanced optimal transport. \textit{J. Meteor. Soc. Japan}, \textbf{102}, 35-47, \url{https://doi.org/10.2151/jmsj.2024-003}.

\bibitem[ECMWF(2024)]{ECMWF2024}
ECMWF, 2024: IFS Documentation. European Centre for Medium-Range Weather Forecasts. [Available at \url{https://www.ecmwf.int/en/publications/ifs-documentation}.]

\bibitem[Gopalakrishnan et al.(2011)]{Gopalakrishnan2011}
Gopalakrishnan, S. G., F. Marks Jr., X. Zhang, J.-W. Bao, K.-S. Yeh, and R. Atlas, 2011: The experimental HWRF system: A study on the influence of horizontal resolution on the structure and intensity changes in tropical cyclones using an idealized framework. \textit{Mon. Wea. Rev.}, \textbf{139}, 1762–1784.

\bibitem[Hagelin et al.(2017)]{Hagelin2017}
Hagelin, S., J. Son, R. Swinbank, A. McCabe, N. Roberts, and W. Tennant, 2017: The Met Office convective-scale ensemble, MOGREPS-UK. \textit{Quart. J. Roy. Meteor. Soc.}, \textbf{143}, 2846–2861.

\bibitem[Hamill et al.(2017)]{Hamill2017}
Hamill, T. M., E. Engle, D. Myrick, M. Peroutka, C. Finan, and M. Scheuerer, 2017: The U.S. National Blend of Models for Statistical Postprocessing of Probability of Precipitation and Deterministic Precipitation Amount. \textit{Mon. Wea. Rev.}, \textbf{145}, 3441–3463.

\bibitem[Han et al.(2024)]{Han2024}
Han T., S. Guo, F. Ling, K. Chen, J. Gong, J. Luo, J. Gu, K. Dai, W. Ouyang, L. Bai, 2024: FengWu-GHR: Learning the Kilometer-scale Medium-range Global Weather Forecasting. \textit{arXiv preprint arXiv:2402.00059.}

\bibitem[Ho et al.(2020)]{Ho2020}
Ho, J., A. Jain, and P. Abbeel, 2020: Denoising Diffusion Probabilistic Models. \textit{arXiv preprint arXiv:2006.11239.}

\bibitem[ICAO(2016)]{ICAO2016}
ICAO, 2016: Guidance on the harmonized WAFS grids for Cumulonimbus cloud, icing and turbulence forecasts (version 2.6). International Civil Aviation Organization, 16pp. [Available at \url{https://www.icao.int/airnavigation/METP/MOG\%20WAFS\%20Reference\%20Documents/WAFS\_HazardGridUserGuide.pdf}.]

\bibitem[Inverarity et al.(2023)]{Inverarity2023}
Inverarity, G. W., W. J. Tennant, L. Anton, N. E. Bowler, A. M. Clayton, M. Jardak, A. C. Lorence, F. Rawlins, S. A. Thompson, M. S. Thurlow, D. N. Walters, and M. A. Wlasak, 2023: Met Office MOGREPS-G initialisation using an ensemble of hybrid four-dimensional ensemble variational (En-4DEnVar) data assimilations. \textit{Quart. J. Roy. Meteor. Soc.}, \textbf{149}, 1138–1164.

\bibitem[JMA(2018)]{JMA2018}
JMA, 2018: Instruction for guidance. \textit{Report of Numerical Prediction Division}, \textbf{64}, 248 pp (in Japanese). [Available at \url{https://www.jma.go.jp/jma/kishou/books/nwpreport/64/No64\_all.pdf}.]

\bibitem[JMA(2022)]{JMA2022}
JMA, 2022: \textit{Annual Report on the Activities of the RSMC Tokyo~-~Typhoon Center 2022.} Japan Meteorological Agency, 143pp. [Available at \url{https://www.jma.go.jp/jma/jma-eng/jma-center/rsmc-hp-pub-eg/AnnualReport/2022/Text/Text2022.pdf}.]

\bibitem[JMA(2024)]{JMA2024}
JMA, 2024: \textit{Outline of the operational numerical weather prediction at the Japan Meteorological Agency.} Japan Meteorological Agency, 262pp. [Available at \url{https://www.jma.go.jp/jma/jma-eng/jma-center/nwp/outline-latest-nwp/index.htm}.]

\bibitem[Kingma and Ba(2014)]{Kingma2014}
Kingma, D. P., and J. Ba, 2014: Adam: A Method for Stochastic Optimization. \textit{arXiv preprint arXiv:1412.6980.}

\bibitem[Lam et al.(2022)]{Lam2022}
Lam, R., A. Sanchez-Gonzalez, M. Willson, P. Wirnsberger, M. Fortunato, F. Alet, S. Ravuri, T. Ewalds, Z. Eaton-Rosen, W. Hu, A. Merose, S. Hoyer, G. Holland, O. Vinyals, J. Stott, A. Pritzel, S. Mohamed and P. Battaglia, 2022: GraphCast: Learning skillful medium-range global weather forecasting. \textit{arXiv preprint arXiv:2212.12794.}

\bibitem[Lang et al.(2024)]{Lang2024}
Lang, S., M. Alexe, M. Chantry, J. Dramsch, F. Pinault, B. Raoult, M. C. A. Clare, C. Lessig, M. Maier-Gerber, L. Magnusson, Z. B. Bouall\'{e}ue, A. P. Nemesio, P. D. Dueben, A. Brown, F. Pappenberger, F. Rabier, 2024: AIFS — ECMWF's data-driven forecasting system. \textit{arXiv preprint arXiv:2406.01465.}

\bibitem[LeCun et al.(2015)]{LeCun2015}
LeCun, Y., Y. Bengio, and G. Hinton, 2015: Deep learning. \textit{Nature}, \textbf{521}, 436–444.

\bibitem[Le Coz et al.(2023)]{LeCoz2023}
Le Coz, C., A. Tantet, R. Flamary, and R. Plougonven, 2023: A barycenter-based approach for the multi-model ensembling of subseasonal forecasts. \textit{arXiv preprint arXiv:2310:17933.}

\bibitem[McCann(1997)]{McCann1997}
McCann, R. J., 1997: A Convexity Principle for Interacting Gases. Advances in Mathematics, \textbf{128(1)}, 153–179.

\bibitem[Nair and Hinton(2010)]{Nair2010}
Nair, V., and G. E. Hinton, 2010: Rectified linear units improve restricted Boltzmann machines. Proceedings of the Twenty-seventh International Conference on Machine Learning (ICML-10), Haifa, Israel, 807–814.

\bibitem[NCEP(2016)]{NCEP2016}
NCEP, 2016: \textit{Global Forecast System~-~Global Spectral Model (GSM)~-~v13.0.2.} [Available at \url{https://vlab.noaa.gov/ web/gfs/documentation}.]

\bibitem[Nishimura and Fukuda(2019)]{Nishimura2019}
Nishimura, S., and J. Fukuda, 2019: Advancement of Tropical Cyclone Track Forecasts. \textit{Yohou Gijutsu Kenshu Text}, \textbf{24}, 114–141 (in Japanese). [Available at \url{https://www.jma.go.jp/jma/kishou/books/yohkens/24/all.pdf}.]

\bibitem[Pathak et al.(2022)]{Pathak2022}
Pathak, J., S. Subramanian, P. Harrington, S. Raja, A. Chattopadhyay, M. Mardani, T. Kurth, D. Hall, Z. Li, K. Azizzadenesheli, P. Hassanzadeh, K. Kashinath, and A. Anandkumar, 2022: FourCastNet: A Global Data-driven High-resolution Weather Model using Adaptive Fourier Neural Operators. \textit{arXiv preprint arXiv:2202.11214.}

\bibitem[Peyr\'{e} and Cuturi(2020)]{Peyre2020}
Peyr\'{e}, G., and M. Cuturi, 2020: Computational Optimal Transport. \textit{arXiv preprint arXiv: 1803.00567.}

\bibitem[Price et al.(2023)]{Price2023}
Price, I., A. Sanchez-Gonzalez, F. Alet, T. R. Andersson, A. El-Kadi, D. Masters, T. Ewalds, J. Stott, S. Mohamed, P. Battaglia, R. Lam, M. Willson, 2023: GenCast: Diffusion-based ensemble forecasting for medium-range weather. \textit{arXiv preprint arXiv:2312.15796.}

\bibitem[Primo et al.(2024)]{Primo2024}
Primo, C., B. Schulz, S. Lerch, and R. Hess, 2024: Comparison of Model Output Statistics and Neural Networks to Postprocess Wind Gusts. \textit{arXiv preprint arXiv:2401.11896.}

\bibitem[Roberts el al.(2023)]{Roberts2023}
Roberts, N., A. Benjamin, E. Gavin, M. Stephen, R. Fiona, S. Caroline, T. Tomasz, A. Paul, B, Laurence. C. Neil, F. Ben, F. Jonathan, G. Tom, H. Leigh, H. Aaron, H. Katharine, J. Simon, J. Caroline, M. Ken, S. Christopher, S. Michael, W. Bruce, B. Simon, B. Mark, B. Daniel, B. Anna, B. Clare, C. Robert, C. Sean, C. Ric, H. Roger, H. Kathryn, H. Teresa, M. Marion, P. Jon, P. Tim, S. Victoria, S. Eleanor, and W. Mark, 2023: IMPROVER: The New Probabilistic Postprocessing System at the Met Office. \textit{Bull Amer. Meteor. Soc.}, \textbf{104}, E680–E697.

\bibitem[Ronneberger et al.(2015)]{Ronneberger2015}
Ronneberger, O., P. Fischer, and T. Brox, 2015: U-Net: Convolutional Networks for Biomedical Image Segmentation. \textit{arXiv preprint arXiv:1505.04597.}

\bibitem[Simon et al.(2018)]{Simon2018}
Simon, A., A. B. Penny, M. DeMaria, J. L. Franklin, R. J. Pasch, E. N. Rappaport, and D. A. Zelinsky, 2018: A description of the real-time HFIP Corrected Consensus Approach (HCCA) for Tropical Cyclone Track and Intensity Guidance. \textit{Wea. Forecasting}, \textbf{33}, 37–57, \doi{10.1175/WAF-D-17-0068.1}.

\bibitem[Song and Ermon(2019)]{Song2019}
Song, Y. and S. Ermon, 2019: Generative Modeling by Estimating Gradients of the Data Distribution. \textit{arXiv preprint arXiv:1907.05600.}

\bibitem[Vannitsem el al.(2021)]{Vannitsem2021}
Vannitsem, S., J. B. Bremnes, J. Demaeyer, G. Evans, J. Flowerdew, S. Hemri, S. Lerch, N. Roberts, S. Theis, A. Atencia, Z. B. Bouall\'{e}gue, J. Bhend, M. Dabernig, L. D. Cruz, L. Hieta, O. Mestre, L. Moret, I. O. Plenkovic, M. Schmeits, J. Ylh\"{a}isi, 2021: Statistical Postprocessing for Weather Forecasts: Review, Challenges, and Avenues in a Big Data World. \textit{Bull Amer. Meteor. Soc.}, \textbf{102}, E681–E699.

\bibitem[Vaswani et al.(2017)]{Vaswani2017}
Vaswani, A., N. Shazeer, N. Parmar, J. Uszkoreit, L. Jones, A. N. Gomez, L. Kaiser, and I. Polosukhin 2017: Attention Is All You Need, \textit{arXiv preprint arXiv:1706.03762.}

\bibitem[Vislocky and Fritsch(1997)]{Vislocky1997}
Vislocky, R. L., and J. M. Fritsch, 1997: Performance of an advanced MOS system in the 1996–97 National Collegiate Weather Forecasting Contest. \textit{Bull. Amer. Meteor. Soc.}, \textbf{78}, 2851–2857.

\bibitem[WMO(2013)]{WMO2013}
WMO, 2013: Cascading Process to Improve Forecasting and Warning Services. \textit{Bulletin nº}, \textbf{62}, 11–15. [Available at \url{https://public.wmo.int/en/resources/bulletin/cascading-process-improve-forecasting-and-warning-services}.]


\end{thebibliography}

\end{document}